\pdfoutput=1

\documentclass[12pt, a4paper]{article}
\usepackage[american]{babel}

\usepackage{authblk}
\usepackage{graphicx}
\usepackage[left=2cm, right=2cm, top=2cm, bottom=2cm]{geometry}

\usepackage{csquotes}
\usepackage[style=numeric-comp,sorting=none,maxbibnames=3,minbibnames=1,backend=biber]{biblatex}
\usepackage{hyperref}
\usepackage{booktabs}
\usepackage{multirow}
\hypersetup{colorlinks=true, linkcolor=blue, citecolor=blue,urlcolor=blue}

\usepackage{algorithm}
\usepackage{algpseudocode}
\usepackage{subcaption}
\usepackage{pifont}
\newcommand{\cmark}{\ding{51}}%
\newcommand{\xmark}{\ding{55}}%

\author[1,4]{Aguirre, Nicolás}
\author[2,3]{Caso, Ramiro}
\author[1,4]{Rodríguez Colmeiro, Ramiro}
\author[2,3]{Santelli, Mauro}
\author[1,2,3]{Toranzo Calderón, Joaquín}

\affil[1]{GIAR, UTN-FRBA}
\affil[ ]{\textit {\{naguirre, rrodriguezcolmeiro, jtoranzocalderon\}@frba.utn.edu.ar}}
\affil[2]{IIF--SADAF--CONICET}
\affil[ ]{\textit {caso.ramiro@conicet.gov.ar}}
\affil[3]{Universidad de Buenos Aires}
\affil[ ]{\textit {mesantelli@uba.ar}}
\affil[4]{Pnyx AI}

\title{A-VERT\\Agnostic Verification with Embedding Ranking Targets}
\date{\today}

\begin{document}
\maketitle

\begin{abstract}
The automatic evaluation of Language Model (LM) responses is a critical piece in the development of benchmarks and metrics, both for model training and quality assessment of production model endpoints. The current approaches to response classification relies on methods that are too expensive (i.e. LLM-as-a-Judge) or that are far from real-world conditions (string-matching, logprob). In this paper, a structure-free evaluation method is presented. The method makes use of semantic embedding distances to match target candidates with arbitrary LM-generated text, resulting in a robust classification of the response at a relatively low compute cost (embedding models of less than $10B$ parameters). The results show a regression score of $~0.97$ and an accuracy of $~96\%$ against human annotators, tested over 3 data sets and 3 different LM architectures.
\end{abstract}

\section{Introduction}\label{sec:introduction}
The current advances in language model (LM) capabilities have triggered more interest in the correct measurement of these models, evidenced by the rapid growth of the literature on new datasets and benchmarks for testing them. Benchmark developers evaluate those capabilities in specific scenarios or tasks such as question-answering (QA)~\cite{westonAICompleteQuestionAnswering2015,suzgunChallengingBIGBenchTasks2022a}, common-knowledge~\cite{hendrycksMeasuringMassiveMultitask2021b,wangMMLUProMoreRobust2024}, coding~\cite{chenEvaluatingLargeLanguage2021a}, translation~\cite{thellmannMultilingualLLMEvaluation2024}, or even reasoning tests~\cite{cholletMeasureIntelligence2019a}, among others.\footnote{A large compilation of benchmarks can be found in~\cite{liangHolisticEvaluationLanguage2023}.} 
In this regard, there is also a wide variety of methods to determine whether an instance of a task has been correctly solved by an LM. 
For example, the method for determining whether a coding task has been satisfactorily completed (where a problem is considered solved when it passes the unit tests and its \texttt{pass@k} metric is reported~\cite{kulalSPoCSearchbasedPseudocode2019}) is not the same as the method for a translation task or a QA task. 
While the former focuses on the result of the code generated by the LM, and not on the content of the code itself, the latter focuses on its content, and it is common to use a criterion of similarity between the target and the generated text. 

In this paper, we focus on the methodology used to determine whether an LM response to a QA task is correct or incorrect. QA tasks are intended to capture a simple question prompt and an answer completion dynamic. They are an extension of the idea of a test in the case of human students, where teachers probe the students' knowledge of a topic to ascertain their competence in an area. In general, the correct resolutions are previously known by teachers---or in our case, benchmark designers.

We can distinguish between ``open-book'' and ``closed-book'' questions, depending on wheth\-er the answer, or the information required to provide an answer, is present in the question or the context that may accompany it. Open-book questions are often aimed at skills that involve reading comprehension or inferring implicit information. Closed-book questions, in contrast, may be seen as a way of making explicit what a model learned during its training, as part of its internal parameters.

Together with the question and a context, a QA task can provide more than one answer, as in a multiple choice question-answering test (MCQA), as long as one of them is correct or preferable over the others. In fact, many benchmarks consist of multiple choice QA tasks,\footnote{See, for instance, the benchmarks MMLU~\cite{hendrycksMeasuringMassiveMultitask2021b} and MMLU-Pro~\cite{wangMMLUProMoreRobust2024}.} since they are built from standardized tests with this format.\footnote{For instance, AR-SAT~\cite{zhongAnalyticalReasoningText2022} is based on the \textit{Law School Admission Test} (LSAT) performed between 1991 and 2016 in the USA.} However, other benchmarks introduce questions without candidate answers. In certain cases, the question itself does not need it, as in a polar question that can be answered either affirmatively or negatively, or in questions that mention relevant alternatives. In other cases, the question offers no clue regarding the correct answer.\footnote{As an example, consider the benchmark bAbI~\cite{westonAICompleteQuestionAnswering2015}, it covers a set of tasks which are all stated as QA tasks with no explicit options. Some of these QA tasks are polar questions and some are WH-questions.}

When prompted with a QA task, an LM will generate a response, but, in principle, there is no predetermined format the response will have. The evaluation of the response can depend heavily on that format, so different benchmark designers have developed some strategies to make evaluation easier. Designing MCQA tests is already a way of inducing an LM to answering in a suitable format, but we can induce this bias with some previous examples in the prompt (``few-shots'') or with some instructions that shape the answer, be it general (as in ``chain-of-thought'' techniques) or precise for the faced QA task (as in ``system prompts'' techniques). However, it is always possible that the LM produces a correct answer that differs from the answer that the benchmark designer posited as a reference (explicitly, as in MCQA tests, or not). Another drawback of these scenarios is that they are hardly similar to the everyday use of an LM, where the user presents a question with limited context, no options, and in a chat-style, expecting a long-form response.

When it comes to evaluating the answer, each of these strategies carries its own problems. MCQA tests facilitate options that could prevent an LM from generating wrong answers by inducing the selection of one of the alternatives presented in the question; not including such alternatives demands that the LM generate an answer relying on its own resources. \emph{Few-shot} tests are similar to MCQA tests in that they induce a vocabulary and grammar (those in the given examples), and again, they provide more clues to the LM under evaluation. Working with \emph{system prompts} is also expected to induce the format of the target answer, but without the need to mention the vocabulary and structure of the target answer;\footnote{It can be done, for instance, to leverage the performance of the LM in the task.} however, a good benchmark should evaluate a set of capabilities without adding more elements to the tests, as is the case with the instructions added in \emph{system prompts}. Finally, \emph{Chain-of-thought} is not meant to induce the format of the answer, but to improve performance on tasks that can benefit from a careful line of reasoning. Thus, CoT tends to produce text that includes, besides an answer to the question, further elements that need to be sorted out.

These expected differences between generated and target answers demand a way of assessing the correctness of the output of a test. This, in turn, can be done in many ways. One way is to manually check each answer and grade it. This, however, is obviously unfeasible for benchmarks that have a huge volume of tests. A solution is to automatically compare generated and target answers. The most straightforward way of comparing answers is \emph{exact match}, a character-by-character string comparison. But there are other ways, each with its own problems. We may divide them into two broad classes, depending on whether we take into account either the expected output or the generated output.

The first class requires computing the probability of generating the target answer and, if available, incorrect alternatives. Methods in this class include \emph{likelihood}, \emph{surprise}, and \emph{perplexity}, among others. Correctness may then be determined by accepting an answer when its value in the relevant measure is lower or higher than a specified threshold (depending on the particulars of the probability measure being used). The selection of a proper threshold is a problem in itself, since there is no obvious choice. If there are incorrect answers that can be measured as well, they may be used as a threshold in the following way: the target answer may be accepted just in case its value is above, or below (depending on the particular probabilistic measure deployed), the values for every incorrect alternative. Notice that for MCQA tests, the method should consider both the complete form of the answers and the labels for each answer (typically a number or letter). While a priori this simplifies the difficulties of having a matching function and a threshold, this method only works if access to the internal parameters of the models is available, which might not hold true for some proprietary models. Furthermore, the \emph{likelihood} method (for instance) considers only a single instance of a correct answer as opposed to free LM generation. This is problematic as there are many ways to answer correctly, and the \emph{likelihood} approach fails to capture this diversity of valid responses. 
Finally, for Reasoning-LM that use \emph{chain-of-thought} and \emph{tool-usage} techniques, this method does not allow reasoning traces to be generated nor tools to be used, making the method incompatible with agent testing.

The second class of methods involves linguistic comparisons between the generated answer and the target. \emph{Exact match}, mentioned above, is a syntactical comparison, since it only takes into account the form of representation. \emph{Exact match} is quite easy to compute and is broadly used in benchmarks for measuring the success of a model in a QA task. Unfortunately, it has a major problem: it is too sensitive to the way in which the answer is phrased, hence prone to false negatives.\footnote{For instance, if we have a target answer $A$, but an LM answers ``The answer to this question is $A$,'' \emph{exact match} would deem it as an incorrect one, which is not the case.}
Other methods apply a function that compares the common syntactical elements of both the generated and the target answer, and the elements present exclusively in each of them. Some of these methods consider the proportion of shared tokens, characters, words, or \emph{n}-grams, over either the total or the disjoint set of elements.\footnote{BLEU~\cite{postCallClarityReporting2018} and ROUGE~\cite{linROUGEPackageAutomatic2004} are two well-known methods of this kind. More methods that capture the guidelines of these ideas can be found in~\cite{gomez-adornoEvaluationSimilarityMeasures2020}.}
These methods accept an answer as correct when its value is above a certain threshold, showing closeness to the target. These methods exhibit certain shared shortcomings: selecting a non-arbitrary threshold, and the possibility of both false positives and negatives, since minor differences (in terms of the selected measure) between generated and target answers could hide non-trivial differences in meaning.\footnote{As an example, consider a generated answer that differs from the target only in a negated sentence or in a word with or without a negative prefix. In both cases, it is likely that both answers result in a high value of syntactical closeness.}
In addition, every method in this syntactical comparison subclass seems to be inadequate for models prompted with instructions prone to long answers, as in \textit{chain-of-thought}, since correct answers will result in low values of syntactical similarity.

As opposed to syntactical comparisons, semantical comparisons intend to deal with the semantical content of the answers, regardless of, for instance, their format or length. These methods include the use of embedding models (see Section~\ref{sec:related}) for computing the similarity between the generated answer and the target. If we only check the similarity between the generated answer and the target, we face the problem of choosing a non-arbitrary threshold. If explicit alternatives to the target are available, we can avoid the threshold problem by computing the similarity of the generated answer both with respect to those alternatives and with respect to the target itself, and then accept the generated answer if it is more similar to the target than to the alternatives. This strategy seems to be underexplored in model evaluation. 

\begin{figure}[H]
  \centering
  \includegraphics[width=0.99\linewidth]{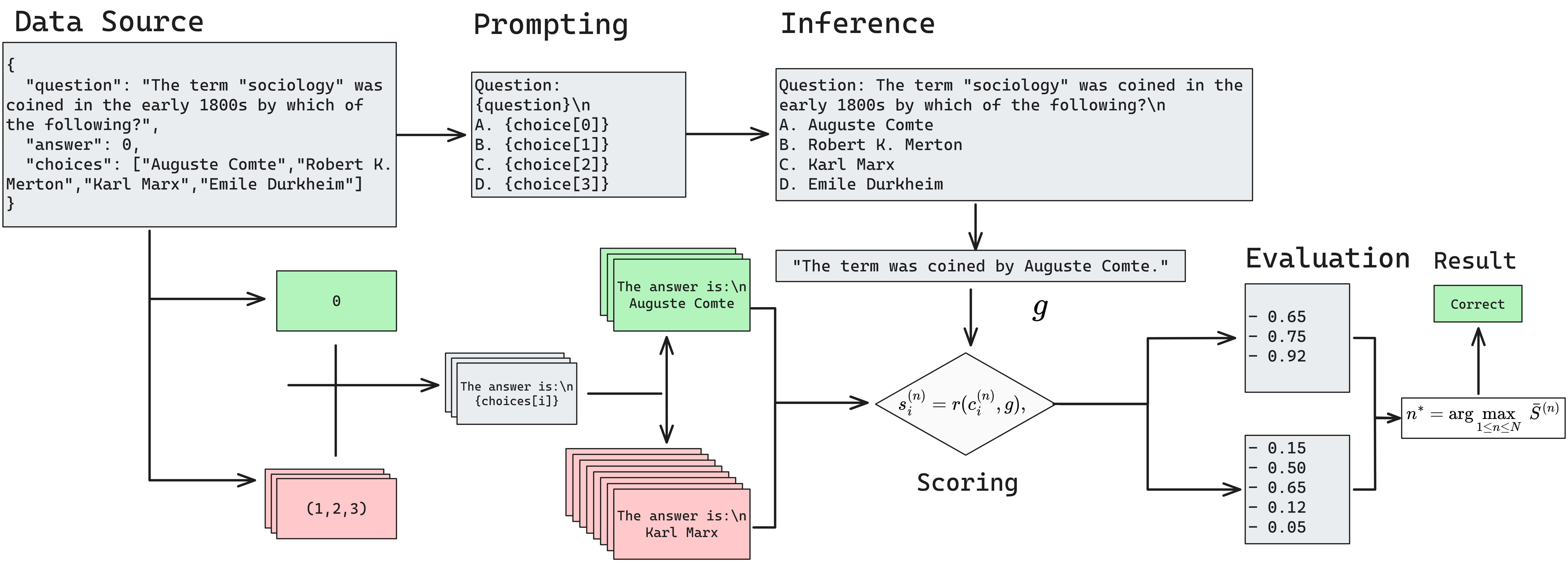}
  \caption{Overview of the \emph{A-VERT} methodology. The process begins with a data source containing question, answer, and choices. During prompting, the question and choices are formatted and presented to the LM. The LM generates a free-form response $g$ during inference. In the evaluation phase, \emph{A-VERT} constructs semantic target groups: a \emph{correct} (green) and \emph{wrong} (red) groups containing enhanced versions of the target answer and incorrect choices, respectively. Each candidate $c_i^{(n)}$ in both groups is scored against the generated response $g$ using a ranking function $r(\cdot)$. Finally, the method selects the representative score for each group, and assigns the LM response to the group with higher ranking score.}
  \label{fig:avert}
\end{figure}

We believe that due to these problems, current approaches to benchmark response evaluation are flawed and need to be improved to allow LMs to provide responses in environments that resemble the real world. Moreover, without the burden of having to adhere to strict generation strategies, benchmark generation can focus on more specific and diverse datasets, rather than being over-reliant on the adoption of frameworks for standardized tests. 
For these reasons, we propose \emph{A-VERT}, an agnostic response verification technique based on ranking different response targets that uses their embeddings as feature, providing a mathematically-based and versatile metric that can be applied to new and old benchmarks. 
The proposed methodology is summarized in Figure~\ref{fig:avert}. 

This paper is structured as follows: Section~\ref{sec:related} reviews the existing literature on syntactical (or lexical) and semantic textual similarity (STS) metrics and evaluation methodologies for LM responses. 
Section~\ref{sec:methodology} presents the mathematical formulation of the \emph{A-VERT} method, detailing both embedding-based and reranker-based ranking functions, along with the experimental setup encompassing benchmark selection, LMs configurations, and evaluation protocols. In terms of reproducibility, the code and the data are publicly available.\footnote{\url{https://github.com/pnyxai/a-vert}}
Section~\ref{sec:results} reports the results of the experiments on multiple benchmarks, demonstrating the performance characteristics of the proposed method. 
Section~\ref{sec:discussion} provides a comprehensive analysis of ranking model performance, ablation studies examining the impact of different scores and sub-tasks, comparative evaluation against traditional benchmark methodologies, and closes with a discussion of the limitations of the approach. Finally, Section~\ref{sec:conclusion} synthesizes key contributions and outlines promising directions for future research in agnostic response verification. 

\section{Related Work}\label{sec:related}

Our research did not find in the literature any technique that addressed the assessment of the free-form response of an LM by comparing classes of correct and incorrect answers. The only techniques found in the literature that compare those classes are based on likelihood and related measures, and hence cannot be applied to the evaluation of free-form responses. We found, however, several works that build up to the construction of the proposed method. In what follows, we report some of the methods used to compare candidate and target answer, which are useful in determining whether the candidate answer is correct, but that have not been used to perform semantic or syntactic comparisons between the candidate answer and the classes of correct and incorrect answers.

\newcommand{\vref}{\mathbf{x}}
\newcommand{\vhyp}{\mathbf{\hat{x}}}
\newcommand{\sentref}{x}
\newcommand{\senthyp}{\hat{x}}
\newcommand{\idf}{{\rm idf}}

\paragraph{BERTscore.}~\cite{zhangBERTScoreEvaluatingText2020} uses contextual embeddings (like those produced by BERT) to represent tokens in target answer $\sentref{} = \langle \sentref{}_1, \dots, \sentref{}_k \rangle$ and candidate sentence $\senthyp{} = \langle \senthyp{}_1, \dots, \senthyp{}_m \rangle$, capturing context-dependent meanings, unlike static word embeddings. It computes cosine similarity (simplified to inner product $\vref_i^\top\vhyp_j$ with prenormalized vectors) between token embeddings from reference and candidate sentences, $\sentref_i$, and $\senthyp_j$, respectively. The method employs greedy matching to align tokens, calculating precision, recall, and F1 scores, considering also the importance of rare tokens by using the inverse document frequency $\idf(w)$. In this way, each token in one sentence is matched to its most similar token in the other sentence based on embedding similarity, enabling soft semantic matching rather than exact syntactical matching.

\paragraph{Semantic answer similarity.}~\cite{rischSemanticAnswerSimilarity2021} addresses the limitations of lexical-based evaluation metrics in QA systems by proposing a cross-encoder-based approach to measure STS between LM responses and ground truth answers. 
The authors trained a cross-encoder model in which the answer pairs are concatenated with a special separator token and processed monotonically through a transformer-based language model. 
Despite the fact that in \cite{rischSemanticAnswerSimilarity2021} the authors use an architecture commonly used for rerankers, unlike the method proposed in this work, they only consider the \emph{correct} category and do not consider \emph{wrong} responses, which can lead to classification errors when the LM response is semantically similar to the target response but incorrect.

\paragraph{SemScore.}~\cite{aynetdinovSemScoreAutomatedEvaluation2024} proposes an evaluation metric for instruction-tuned LMs based on STS. The method operates in two steps: (1) it separately embeds both the model response and the target response using an embedding model; (2) it computes the cosine similarity between the resulting sentence-level embeddings to produce the final SemScore value. 
Unlike token-level approaches such as BERTScore, SemScore operates at the sentence level, directly comparing semantic representations of entire responses rather than performing token-by-token alignment. The resulting similarity between the target response and the LM response score lies within $[-1, 1]$, where values closer to $1$ indicate semantically similar sequences. To evaluate their proposed metric, the authors tagged LM responses with human annotation in a four-category rating system, from more to less acceptable responses. Then, the average of the human annotation was used as a gold standard for the similarity between human and LM responses. In that work, the authors do not consider the use of this metric to match correct responses, limiting themselves only to extracting a real-valued metric.

\paragraph{LLM-as-a-Judge.}~\cite{zhengJudgingLLMasaJudgeMTBench2023} addresses the challenge of evaluating LM-based chat assistants on open-ended questions using strong LLMs as automated judges to approximate human preferences. 
The method compares three evaluation approaches: (1) pairwise comparison, where an LLM judge determines which of two responses is better; (2) single-answer grading, where a score is directly assigned to individual responses; and (3) reference-guided grading, which incorporates reference solutions when available. 
The authors claim that GPT-4 as a judge achieved over 80\% agreement with human evaluators, matching the level of human-human agreement. 
However, the authors identified that the approach exhibits several biases, such as position bias (favoring responses in certain positions), verbosity bias (preferring longer responses), and self-enhancement bias (potentially favoring responses from the same model family). The authors propose mitigation strategies such as position swapping, chain-of-thought prompting, and reference-guided evaluation to address these limitations.

\section{Methodology}\label{sec:methodology}

The proposed method relies on measuring distances in the high-dimensional space of semantic embedding models and groups of candidate responses that share the same semantic information. This approach makes \emph{A-VERT} agnostic to the types of questions (open-ended or multiple-choice) as long as the question provides a target response (or a group of accepted responses) and a group of incorrect ones. The use of semantic groups to assign the semantic meaning of an LM generation relieves the proposed scoring method from setting manual thresholds that depend on the context of the benchmark under study. Moreover, the method relies on standard models to provide STS between the candidate groups and the target group; such models can be replaced as the state of the art improves without having to modify the \emph{A-VERT} implementation.

\subsection{A-VERT}
Formally, the \emph{A-VERT} method, presented in Figure~\ref{fig:avert}, can be described as follows. Given a generated string $g$ and $N$ target groups of semantically similar strings, each containing $I_n$ candidates $c_i^{(n)}$:

\begin{equation}\label{eq:candidate_gruop}
    \{\, c_i^{(n)} \mid i = 1, \dots, I_n \,\},
\end{equation}

\noindent{}for each candidate $c_i^{(n)}$, a semantic rank is produced by:

\begin{equation}
    \label{equ:ranker_fn}
    s_i^{(n)} = r(c_i^{(n)}, g),
\end{equation}

\noindent{}where $r(\cdot)$ is a ranking function. Then, for each group $n$, a representative is elected:

\begin{equation}
    S^{(n)} = \max_{1 \leq i \leq I_n} \; r\!\left(s_i^{(n)}\right),
\end{equation}

\noindent{}and all representative ranking scores $S^{(n)}$ are normalized:

\begin{equation}
    \bar{S}^{(n)} = \frac{S^{(n)}}{\sum_{n=1}^{N} S^{(n)}}.
\end{equation}

\noindent{}Finally, the selected group $n^\ast$ is obtained as:

\begin{equation}\label{eq:selec_group}
    n^\ast = \arg\max_{1 \leq n \leq N} \; \bar{S}^{(n)}.
\end{equation}

\noindent{}The complete target selection procedure of the \emph{A-VERT} method is described in Algorithm~\ref{algo:avert}.

\begin{algorithm}
\caption{A-VERT target selection}
\label{algo:avert}
\begin{algorithmic}[1]
\State \textbf{Input:} $N$ groups $\{c_i^{(n)}\}_{i=1}^{I_n}$, generated string $g$, ranking function $r(\cdot)$
\For{$n \gets 1$ to $N$}
    \State $S^{(n)} \gets \max\limits_{1 \leq i \leq I_n} r(c_i^{(n)}, g)$
\EndFor
\For{$n \gets 1$ to $N$}
    \State $\bar{S}^{(n)} \gets S^{(n)} / \sum_{n=1}^{N} S^{(n)}$
\EndFor
\State $n^\ast \gets \arg\max\limits_{1 \leq n \leq N} \bar{S}^{(n)}$
\State \textbf{Output:} $n^\ast$
\end{algorithmic}
\end{algorithm}

The method \emph{A-VERT} relies on the accuracy of $r(\cdot)$, which is responsible for producing a metric whose value is higher when the STS of $g$ and $c$ are close. The main candidates for $r(\cdot)$ are embedding-based distance metrics and reranker models.

\paragraph{Embedding-Based Ranks.}
Embedding models are a type of machine learning model oriented at capturing the semantic meaning and contextual relationships of texts, be they sentences or complete documents. Their goal is to produce a dense numerical vector representation, known as an embedding. In this embedding, semantically similar texts are closer together and separate from those with different semantic content. This distance is normally measured using a cosine similarity metric or a dot product. Not long ago, embedding models experienced a significant improvement with the development of the bidirectional encoder-type transformer architecture, such as BERT. However, more current developers are approaching this technique based on fine-trained causal LMs, trying to take advantage of the high expressivity of the base model. These models are normally trained using contrastive learning. This type of training minimizes the distance of embeddings in positive pairs (semantically similar texts) and maximizes the distance between negative pairs (semantically dissimilar texts). The resulting embedding is then forced to create representations where similar concepts are close and dissimilar ones are apart.

The implementation of these models into a ranking function is then straightforward, as they are trained for the task of interest in \emph{A-VERT}. Thus, a ranking function can be built using the cosine distance and an embedding model $\mathcal{E}(\cdot)$ by:

\begin{equation}
    r_{e}(c, g) = 1 - \frac{\mathcal{E}(c) \cdot \mathcal{E}(g)}{\|\mathcal{E}(c)\|_2\ \|\mathcal{E}(g)\|_2}.
\end{equation}

\paragraph{Reranker-Based Ranks.}
Reranker models are commonly used to refine vector database searches by taking a pair of texts (normally called \emph{query} and \emph{document}) and producing a \emph{relevance score} between them. The topology of the model is similar to the embeddings, built upon bidirectional or causal transformer blocks. The training process of these models is done in a supervised manner and often as a fine-tuning of a base embedding model. During training, the model is presented with a document pair of texts and a target score, 0 for irrelevance and 1 for strong relevance. The resulting model tends to perform better than the embedding model but at a higher computational cost, since the query inputs are longer as both texts must be present in the input and the computational complexity of transformer blocks is quadratic with the query length ($O(N^2)$).

Given that these models are already trained to produce a single relevance score for a document pair, a re-ranked model $\mathcal{R}(\cdot)$ can be implemented directly as a ranking function:

\begin{equation}
    r_{r}(c, g) = \mathcal{R}(c, g).
\end{equation}

\subsubsection{Target Groups Construction}\label{group-contruction}

The target groups used in \emph{A-VERT} are groups of semantic textual similarity. The groups are not limited to any kind in particular, but in this work only two groups are analyzed: the \emph{correct} and the \emph{wrong} groups. These groups have a meaning only when paired with a specific query or question, then the \emph{correct} group will contain texts that have different strings that share the same meaning as the target response. For example, the \emph{correct} group for the question ``what color is the sky?" will contain ``the sky is blue,'' ``it is blue,'' ``blue, on a clear day,'' etc. The \emph{wrong} group then will contain other related texts but that will not be regarded as valid answers to the question, in our example they could be ``The color of the sky is the effect of refraction,'' ``Probably green,'' ``that is a great question, let us ask a meteorologist!,'' etc.

For benchmark questions, the answer comes in mainly two forms, open world and multiple choice, but in either case most benchmarks provide a single string (or even a single character) as target. This poses a difficulty when trying to extract meaning from it since they lack context. While creating a semantically rich target is the job of the dataset creator, some enhancements can be included to any target string to allow for better group construction. To alleviate this problem, the following enhancements are applied to the target strings (either in the \emph{correct} or \emph{wrong} group).

For open-book two enhancements are applied:
\begin{itemize}
    \item \texttt{"The answer is : \{target\} . Let me explain why"}
    \item \texttt{"Therefore, the answer is : \{target\}"}
\end{itemize}
where \texttt{"\{target\}"} is replaced with the candidate response to be added to a group. These two enhancements were created after observing the tendency of LMs to provide richer responses (with explanations) when prompted using chat-templates. 
For multiple-choice questions, we construct the target answer as \texttt{"\{symbol\} : \{target\}"}, where \texttt{"\{symbol\}"} is the symbol assigned to a question option, i.e. a number or a letter. Additionally, multiple-choice questions are enhanced with the following:
\begin{itemize}
    \item \texttt{"Therefore, the correct answer is option \{symbol\}: \{target\}"}
    \item \texttt{"The answer is option \{symbol\}: \{target\}"}
    \item \texttt{"The answer is the \{cardinal\} one, option \{symbol\}. \{target\}. Let me ex\-plain why: " + \{all-choices\}}
    \item \texttt{"Analyzing the options: " + \{all-choices\} + "Therefore, the answer is the \{cardinal\} one, option \{symbol\}. \{target\}"}
\end{itemize}
where \texttt{"\{all-choices\}"} is the list of all choices, using the form: \texttt{"Option \{symbol\}. \{target\}. Is not correct."} or, for the selected option, \texttt{"Option \{symbol\}. \{target\}. Is correct."}

\subsection{Experimental Setup}
The proposed technique is tested using human annotations as the gold standard for a series of benchmarks and open-weights LMs, calculating the agreement between \emph{A-VERT} selected response, [correct, wrong], and the human selected response. In addition, it is analyzed how normal benchmarking approaches affect the retrieved scores for the different datasets and models, and the agreement of these with \emph{A-VERT} and human annotations is also compared.

In the following paragraphs, we will describe the benchmarks, LMs and embedding/reranker models used, followed by a description of the the experimental procedures.

\subsubsection{Benchmarks}
Table \ref{table:benchmark_summary} summarizes the configuration of each benchmark and each scoring method.

\begin{table}[htb!]
\caption{Configuration summary for A-VERT, Logprob and Exact-Match scores.}
\centering
\begin{tabular}{llll}
\toprule
\textbf{Score} & \textbf{Benchmark} &   \textbf{Chat Template} & \textbf{FewShots}\\
\midrule
\multirow{3}{*}{A-VERT} & bAbi & \multirow{3}{*}{ \shortstack[c]{\cmark}} & \multirow{3}{*}{ \shortstack[c]{\xmark}} \\
 & MMLU &  &  \\
 & MMLU-PRO &  &  \\
\cmidrule{2-4}
\multirow{2}{*}{Exact-Match} & bAbi & \xmark & \multirow{3}{*}{ \shortstack[c]{3}} \\
 & MMLU & \cmark &  \\
Logprobs & MMLU-PRO &  \xmark & \\
\bottomrule
\footnotesize{}
\end{tabular}
\label{table:benchmark_summary}
\end{table}

\paragraph{MMLU / MMLU-Pro - Multiple-choice responses.} 
The MMLU and MMLU-Pro are two popular benchmarks that present the LM with a series of multiple-choice questions in a wide set of domains. Both benchmarks possess similar questions, while MMLU has a broader domain variety, MMLU-Pro is smaller, but is presented more often as a ``harder" version of the former. 

The standard measurement procedure of these benchmarks is by means of analyzing the log-probability of each target option and selecting as the LM answer the option with the higher probability. This is done by sending the LM a \emph{completion} (i.e. without associated format, unlike a \emph{chat-completion}) task with a query including three example questions and their responses followed by a fourth question and a selected choice. The LM is not tasked to produce any tokens, only to provide the log-probabilities of generating the last token (normally an option token, like A, B, C, etc.). The LM answer is then extracted from the N queries performed, selecting that with the higher generation probability, if the selected one corresponds to the target, the query is marked as \emph{correct}, if not, it is \emph{wrong}.

In the case of MMLU it is tested using the \emph{generative} variant\footnote{While both MMLU and MMLU-Pro use \emph{logprobs} as the default metric, the generative variant is used here for MMLU to allow an additional method comparison which is closer to real-world scenarios.}, where the LM is prompted using the full chat-template and the examples are given as pairs of user-assistant interactions. In this scenario, the LM response is extracted from the generated string using regex patterns and then an exact-match of the extracted string and the target string is used to assign the \emph{correct} or \emph{wrong} label.

To implement \emph{A-VERT} we allow the LM to freely generate the response to the question. We prompt the LM using its chat-template, provide no few-shots and no system prompts to guide the generation, and collect the model response. This way of querying the LM is closer to the real-world interactions of the models and allows it to use enhancements in computation time that improve their responses~\cite{snellScalingLLMTestTime2024a,zhangChainPreferenceOptimization}. The LM is allowed to generate up to $7000$ tokens (including reasoning traces if available), and if the model fails to return a response given this number of tokens, the query is marked as \emph{invalid} and computed as \emph{wrong}. Then the successful generations of the LM are compared to the target groups using Algorithm~\ref{algo:avert} and a group response is assigned, either \emph{correct} or \emph{wrong}.

The semantic target groups used are two: \emph{correct} and \emph{wrong}. The \emph{correct} group is constructed using the target key (a letter) and the target string (the description of the option). The \emph{wrong} group is constructed in the same way as before, but including all options that are not the target. In addition, target groups are enhanced using a series of ad hoc procedures, we call this process \emph{prompt-enhancement} and it is described in Section~\ref{group-contruction}. 

\paragraph{bAbI - Open-ended responses.} The benchmark bAbI, by~\cite{westonAICompleteQuestionAnswering2015}, proposes a series of questions that aim to measure the basic reasoning capabilities of an LM. This benchmark is selected because it provides an open-ended response type (i.e. not a multiple-choice test) where the model has to generate a response, not select one.

The benchmark proposes a measurement procedure where the LM is prompted using non-zero amount of examples (3 are used in this study) with answers followed by the prompt that is being evaluated. Then the generated string is processed using an ad hoc algorithm that tries to extract the word that follows the ``Answer:'' keyword and finally compares this extraction (strictly) to the target string. If the match is positive, then the answer is assumed to be \emph{correct} if not, it is \emph{wrong}.

In the \emph{A-VERT} implementation, the LM is prompted using a chat template and no system prompt or guidance. In addition, no examples are given before sending the query. The LM is allowed to generate up to $7000$ tokens as output, including any reasoning trace (if the model uses that technique). The generation is then compared with the target string, which consists of the \emph{correct} target and a series of alternate responses that comprise the \emph{wrong} targets. Both groups are enhanced using the process described in~\ref{group-contruction}.
It is important to note that the original benchmark does not provide data to construct the \emph{wrong} targets, and it only provides a single string for the \emph{correct} target (with no context). To generate valid examples of the \emph{wrong} targets, the dataset generation code was analyzed and all entities that can be alternate answers in a given task were evaluated. For example, for a question of \emph{bAbI Task 5}, that poses questions in the form of ``who gave X to Y? Answer: Z" where ``X" is an object and ``Y" and ``Z" are person names, all names that the dataset generation code can use in that task are retrieved and they are assigned to ``Z,'' to create a list of \emph{wrong} targets. Also, using the same example, the targets were enhanced to a contextualized response. Instead of using a single string for a candidate target: ``Z,'' a complete phrase is used: ``Z gave X to Y". This process is purely deterministic based on the task of the dataset.

\subsubsection{Language Models}

Three different open-weight models were selected to evaluate the \emph{A-VERT} method, on the basis of their different creators and architectures. The selected models are the following:

\begin{itemize}
    \item \textbf{Meta Llama 3.3 70B Instruct}~\cite{grattafioriLlama3Herd2024} : Released in December 2024, based on an auto-regressive decoder-only architecture with grouped-query attention. It is pre-trained on public data and instruction tuned using Supervised Fine-Tuning (STF) and Reinforcement Learning with Human Feedback (RL-HF).
    \item \textbf{Qwen3 30B A3B}~\cite{yangQwen3TechnicalReport2025} : Released on April 2025, based on a Mixture of Experts (MoE) topology. Pre-trained on public data and fine tuned for reasoning skills and long context queries. Includes a ``thinking mode" for test-time computing improvements.
    \item \textbf{GPT-OSS 20B}~\cite{openaiGptoss120bGptoss20bModel2025} : Released in August 2025, based on a MoE topology. It is pre-trained using large-scale distillation and fine-tuned using RL and Chain of Thought (CoT).
\end{itemize}

\subsubsection{Embedding \& Reranker Models}

Several embedding and reranker models are analyzed to assume the role of the ranker in Eq.~\ref{equ:ranker_fn}. The selection corresponds to what the authors believe to be a good snapshot of the state-of-the-art in the area. The included models are:

\begin{itemize}
    \item \textbf{Qwen3 Family}~\cite{zhangQwen3EmbeddingAdvancing2025} (6 models): Embedding and reranker models with $0.6B$, $4B$ and $8B$ parameters each. All models with instruction-aware queering capabilities and $32K$ tokens of context length.
    \item \textbf{GTE Family}  (2 models): Modern-BERT based embedding~\cite{warnerSmarterBetterFaster2024} and reranker from Alibaba-NLP, both with $150 M$ parameters and $8192$ tokens of context length.
    \item \textbf{BGE Reranker}~\cite{chenBGEM3EmbeddingMultiLingual2024} (1 model): Reranker from BAAI, version \emph{v2-m3} with $568 M$ parameters and $8192$ tokens of context length.
    \item \textbf{Jina Reranker} (1 model): Reranker from JinaAI, version \emph{v2-base-multilingual} with $278 M$ parameters and $1024$ tokens of context length.
    \item \textbf{Microsoft Multilingual E5}~\cite{wangMultilingualE5Text2024}: (1 model) Embedding model, version \emph{large-instruct}, with $560 M$ parameters, $512$ tokens of context length and instruction aware.    
\end{itemize}

Each model is tested using four different configurations (depending on the model capability):
\begin{enumerate}
    \item \textbf{Base}: The model receives the texts to analyze without any instruction or modification from what is present in the benchmark target.
    \item \textbf{Instruction}: The model receives an instruction to guide the embedding or re-ranking process. The instruction is:\\ \texttt{Instruct: Find the document that best represents the meaning in the que\-ry. Check for any doubts about the question or options. Focus on exact num\-bers, dates, or symbols. Query:\{query\}},\\ where the string \texttt{"\{query\}"} is replaced with LM response.
    \item \textbf{Enhance}: The model receives enhanced targets in addition to the original targets present in the dataset. These are created following the process described in Section~\ref{group-contruction}.
    \item \textbf{Instruction + Enhance}: The model receives both the enhanced queries and these are processed using the provided instruction. This is the full implementation of \emph{A-VERT}.
\end{enumerate}

\subsubsection{Data Acquisition and Human Annotations}
The three LMs are tested on the described benchmarks, using their proposed evaluation for the first $60$ samples in each task (a total of $5460$ samples). Then, for each of the $11$ candidates for ranking models, the \emph{A-VERT} benchmark is also processed for the first $60$ samples in each task for each LM (resulting in $16380$ samples per embedding/reranker model).
All tests are performed using the same procedure with \texttt{lm-eval-harness} library~\cite{sutawikaEleutherAILmevaluationharnessV04912025a}\footnote{The lm-eval tasks used for testing are released in the a-vert repository.}. Since all the tests are performed with the same random seed, the human annotators only need to evaluate the $16380$ samples of a single \emph{A-VERT} test to obtain tags for each of the $11$ embedding/reranker models being tested. The human annotators are asked to rank the LM generation as \emph{correct} if the LM response corresponds to the target label and as \emph{wrong} if the LM response fails to respond as expected (refusals, questioning the prompt as flawed or generation exhaustion are treated as \emph{wrong} tags).

The agreement of the \emph{A-VERT} method and the standard benchmark metrics is compared to the human tags for each of the subtasks in the three datasets: bAbI, MMLU and MMLU-Pro. For the datasets' results in the \emph{A-VERT}, \texttt{Qwen3-Reranker-8B} is used as score model. For bAbI in the standard method a simple completion followed by an extraction and exact-matching technique is used. For MMLU, a full chat-template is used, with three examples formatted as multi-turns (user-assistant interactions), followed by an extraction and exact-matching technique. Finally the MMLU-Pro benchmark uses the accuracy calculation based on the log-probabilities of the target token being generated and basic generation (no chat-format). 

\section{Results}\label{sec:results}

\subsection{Ranking Models and Ablation Tests}

\begin{figure}[H]
    \centering
    \includegraphics[width=0.9\linewidth]{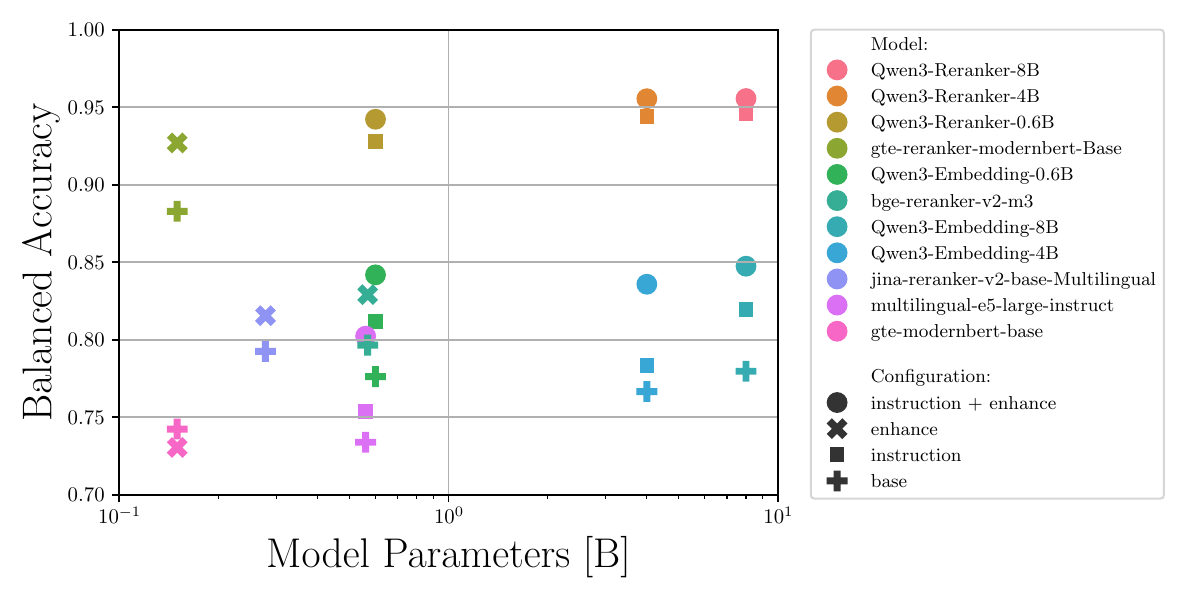}
    \caption{Balanced accuracy of the \emph{A-VERT} method for each model in different settings. The balanced accuracy is calculated over the three benchmarks and three datasets (bAbI, MMLU and MMLU-Pro).}
    \label{fig:all_embeddings}
\end{figure}

For each of the re-ranking and embedding models, and for each configuration (i.e. using instructions or not, using enhancements or not, etc) their balanced accuracy~\cite{mosleyBalancedApproachMulticlass2013} is calculated, using the human annotators as gold standard. The results are shown in Figure~\ref{fig:all_embeddings}, where the models' balanced accuracy (for all tagged samples) is displayed as a function of the number of parameters of each model. It is important to note that all models have less than $10 B$ parameters, which is considered a low count of parameter for models used in \emph{LLM-as-a-Judge} settings. The best scores for each model are presented in Table~\ref{table:all_embeddings}, being \texttt{Qwen3-Reranker-8B} the best one, achieving a balanced accuracy of $0.956$ and a F1 score of $0.986$.

\begin{table}[htbp]
\centering
\caption{Best results for each model's setting in the \emph{A-VERT}. The metrics are calculated over the three benchmarks and three datasets (bAbI, MMLU and MMLU-Pro).}
\resizebox{\textwidth}{!}{%
\begin{tabular}{lllllll}
\toprule
\textbf{Type} & \textbf{Model} &  \textbf{Size (B)} & \textbf{Precision} & \textbf{Recall} & \textbf{F1} & \textbf{Acc*} \\
\midrule
\multirow{6}{*}{ \shortstack[c]{Reranker}}      & Qwen3-Reranker-8B & 8 & 0.984 & \textbf{0.987} & \textbf{0.986} & \textbf{0.956} \\
                                                & Qwen3-Reranker-4B & 4 & \textbf{0.985} & 0.983 & 0.984 & 0.955 \\
                                                & Qwen3-Reranker-0.6B & 0.6 & 0.983 & 0.962 & 0.972 & 0.942 \\
                                                & bge-reranker-v2-m3 & 0.6 & 0.970 & 0.769 & 0.858 & 0.829 \\
                                                & jina-reranker-v2-base-Multilingual & 0.3 & 0.963 & 0.772 & 0.857 & 0.816 \\
                                                & gte-reranker-modernbert-Base & 0.15 & 0.981 & 0.938 & 0.959 & 0.927 \\
\cmidrule{2-7}
\multirow{5}{*}{ \shortstack[c]{Embbeding}}     & Qwen3-Embedding-8B & 8 & 0.969 & 0.819 & 0.888 & 0.847 \\
                                                & Qwen3-Embedding-4B & 4 & 0.967 & 0.798 & 0.874 & 0.836 \\
                                                & Qwen3-Embedding-0.6B & 0.6 & 0.966 & 0.820 & 0.887 & 0.842 \\
                                                & multilingual-e5-large-instruct & 0.6 & 0.966 & 0.726 & 0.829 & 0.802 \\
                                                & gte-modernbert-base & 0.15 & 0.953 & 0.629 & 0.758 & 0.742 \\
\bottomrule
\footnotesize{* Balanced Accuracy}
\end{tabular}%
}
\label{table:all_embeddings}
\end{table}

\subsection{Target Groups Analysis}

\begin{figure}[htb!]
    \centering
    \begin{subfigure}{0.45\textwidth}
        \centering
        \includegraphics[width=0.9\linewidth]{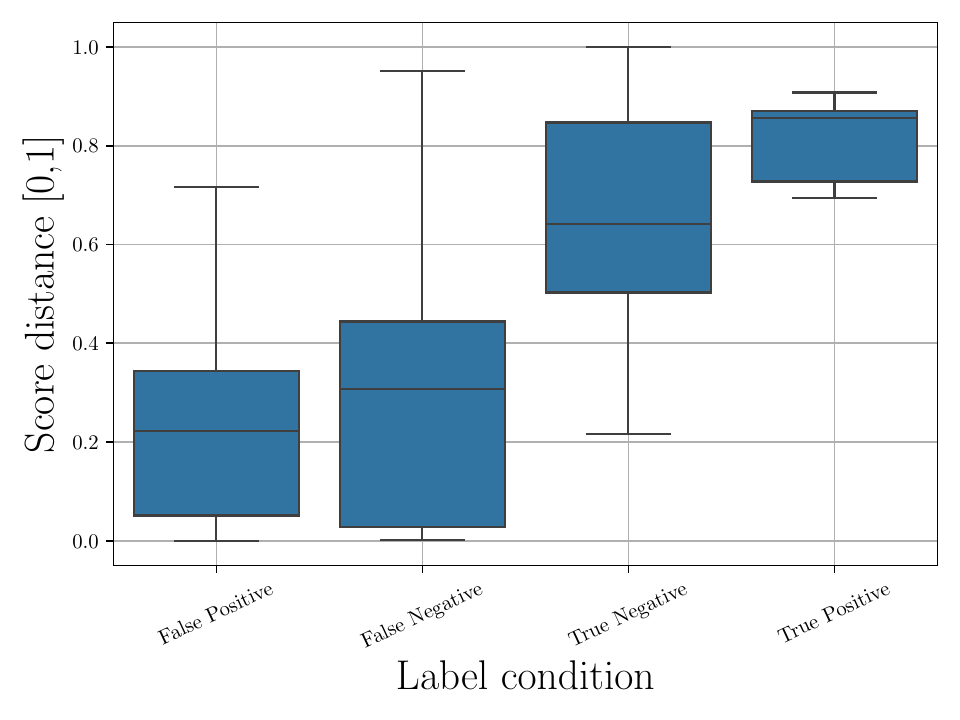}
        \subcaption{Open-ended responses\\(bAbI)}
        \label{fig:babi_dist_targets}
    \end{subfigure}
    \begin{subfigure}{0.45\textwidth}
        \centering
        \includegraphics[width=0.9\linewidth]{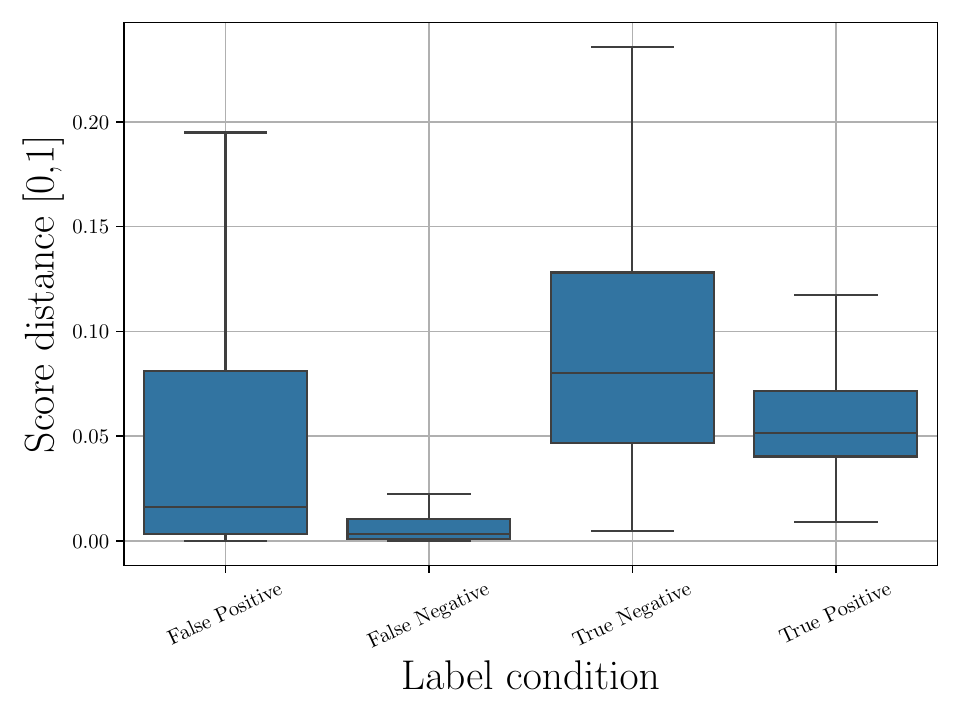}
        \subcaption{Multiple-choice responses\\(MMLU and MMLU-Pro)}
        \label{fig:mmlu_dist_targets}
    \end{subfigure}
\caption{Box-plot of distances between the \emph{A-VERT} scores for the \emph{correct} and \emph{wrong} target groups for each benchmark response type. The distances are grouped in the confusion matrix labels, lower average value in the distance between the target groups signals less confidence in the ranking scores. A value of zero means equal weight to each group (random choice), a score of $1.0$ means total confidence for the selected target group.}
\label{fig:all_dist_targets}
\end{figure}

The raking score separation for the best model, \texttt{Qwen3-Reranker-8B}, is shown in Figure~\ref{fig:all_dist_targets} as one box-plot for each confusion matrix quadrant. A good ranker should assign near zero values to the \emph{False-Positive} and \emph{False-Negative} groups and values near $1.0$ to the \emph{True-Negative} and \emph{True-Positive} groups.

\subsection{Benchmarks Test}

\begin{figure}[htb!]
\centering
    \begin{subfigure}[b]{0.48\textwidth}
        \centering
        \includegraphics[width=\linewidth]{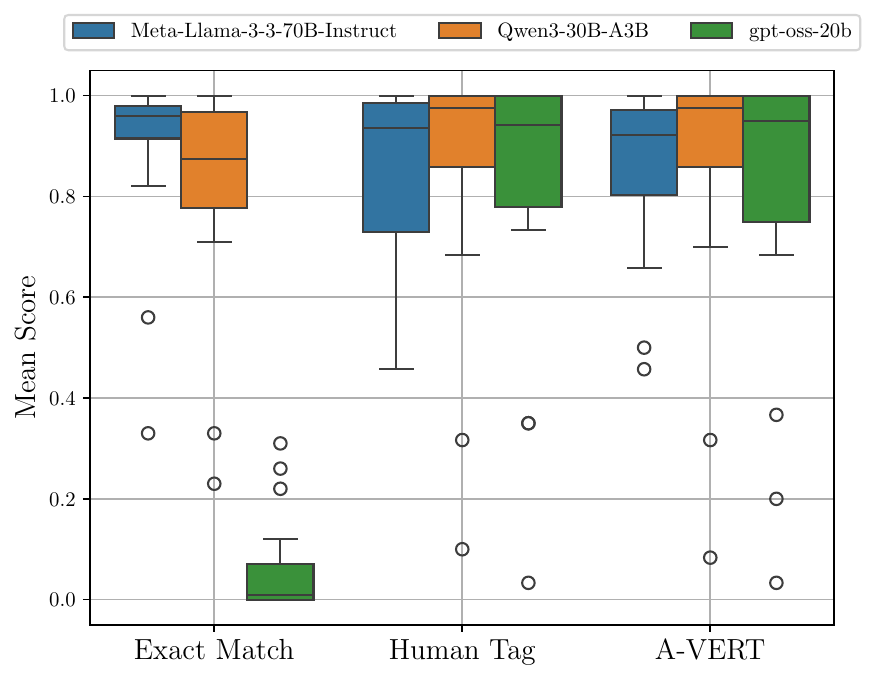}
        \subcaption{}
        \label{fig:babi_comparison}
    \end{subfigure}
    \begin{subfigure}[b]{0.48\textwidth}
        \centering
        \includegraphics[width=\linewidth]{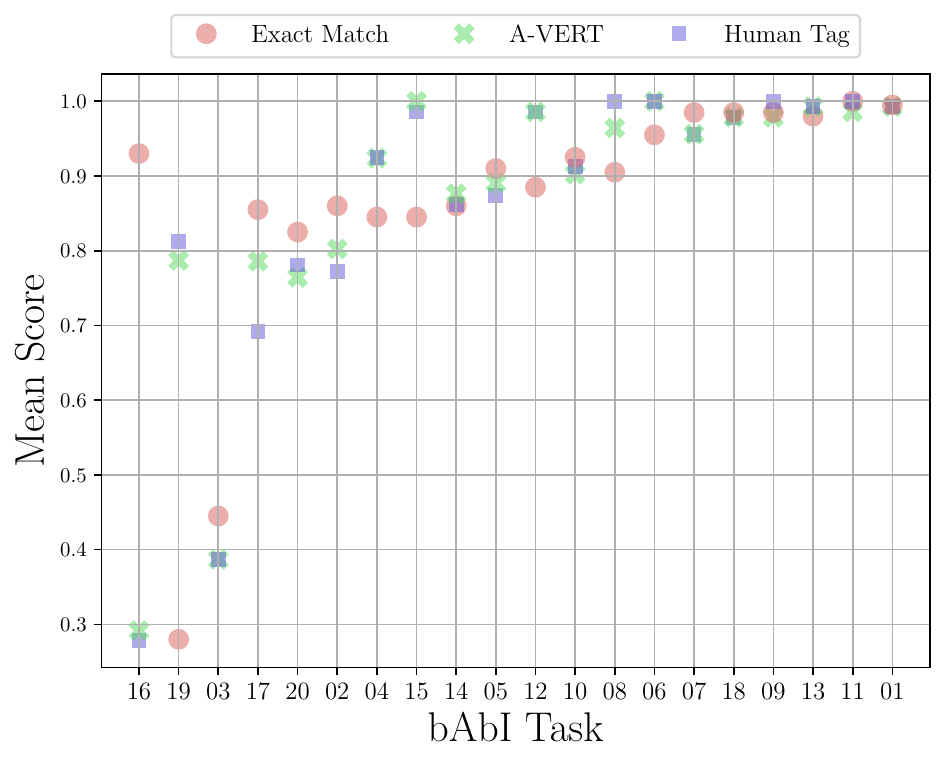}
       \subcaption{}
        \label{fig:babi_comp_task}
    \end{subfigure}
\caption{Comparison of the mean scores in the bAbI task for each of the analyzed methods: exact match, \emph{A-VERT} and the human tag.  \protect\subref*{fig:babi_comparison}) For different LMs: \emph{{GPT-OSS 20B}}, \emph{Meta Llama 3.3 70B Instruct}, and \emph{Qwen3 30B A3B} are in blue, orange, and green, respectively. \protect\subref*{fig:babi_comp_task}) For bAbI subtask scores: exact match, \emph{A-VERT}, and human tag are represented as red circle, green cross and blue square, respectively. The \emph{{GPT-OSS 20B}} results are excluded from this graph to better assess the agreement of the methods.}
\label{fig:babi_comparison_full}
\end{figure}

\begin{figure}[htb!]
\centering
    \begin{subfigure}[b]{0.48\textwidth}
        \centering
        \includegraphics[width=\linewidth]{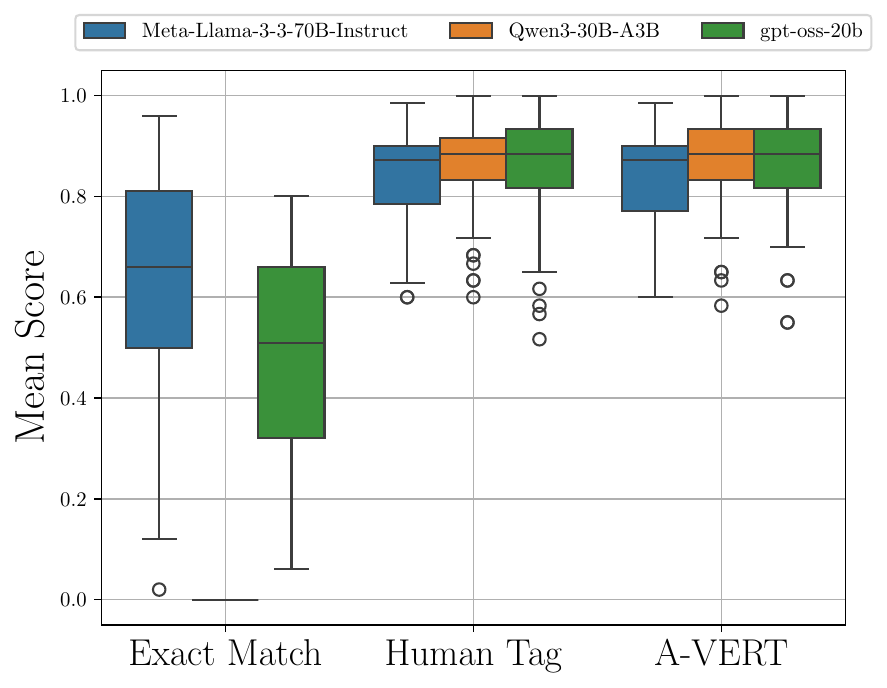}
        \subcaption{}
        \label{fig:mmlu_dist_comparison}
    \end{subfigure}
    \begin{subfigure}[b]{0.48\textwidth}
        \centering
        \includegraphics[width=\linewidth]{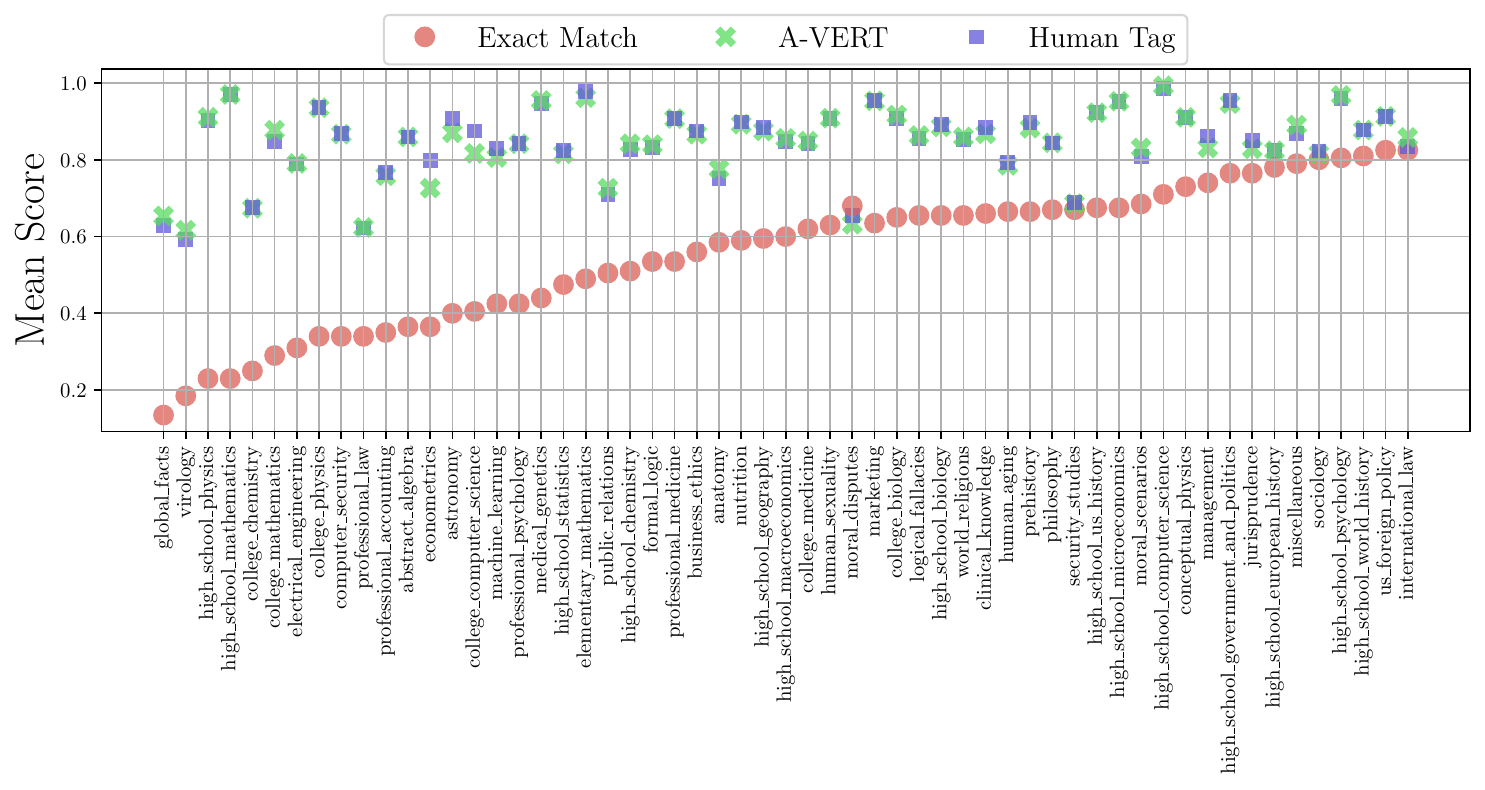}
        \subcaption{}
        \label{fig:mmlu_comparison_task}
    \end{subfigure}
\caption{Comparison of the mean scores in the MMLU task for each of the analyzed methods: exact match, \emph{A-VERT} and the human tag. \protect\subref*{fig:mmlu_dist_comparison}) For different LMs: \emph{{GPT-OSS 20B}}, \emph{Meta Llama 3.3 70B Instruct}, and \emph{Qwen3 30B A3B} are in blue, orange, and green, respectively. \protect\subref*{fig:mmlu_comparison_task}) For MMLU subtask scores: exact match, \emph{A-VERT}, and human tag are represented as red circle, green cross and blue square, respectively. The \emph{{Qwen3 30B A3B}} results are excluded from this graph to better asses the agreement of the methods.}
\label{fig:mmlu_comparison_full}
\end{figure}

\begin{figure}[htb!]
\centering
    \begin{subfigure}[b]{0.48\textwidth}
        \centering
        \includegraphics[width=\linewidth]{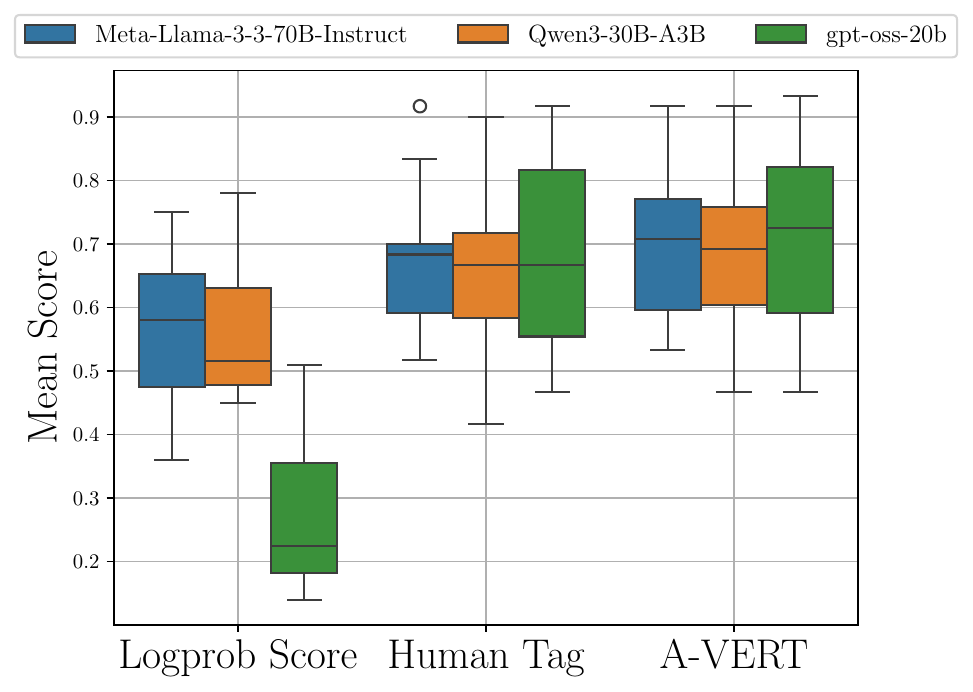}
        \subcaption{}
        \label{fig:mmlu-pro_comparison}
    \end{subfigure}
    \begin{subfigure}[b]{0.48\textwidth}
        \centering
        \includegraphics[width=\linewidth]{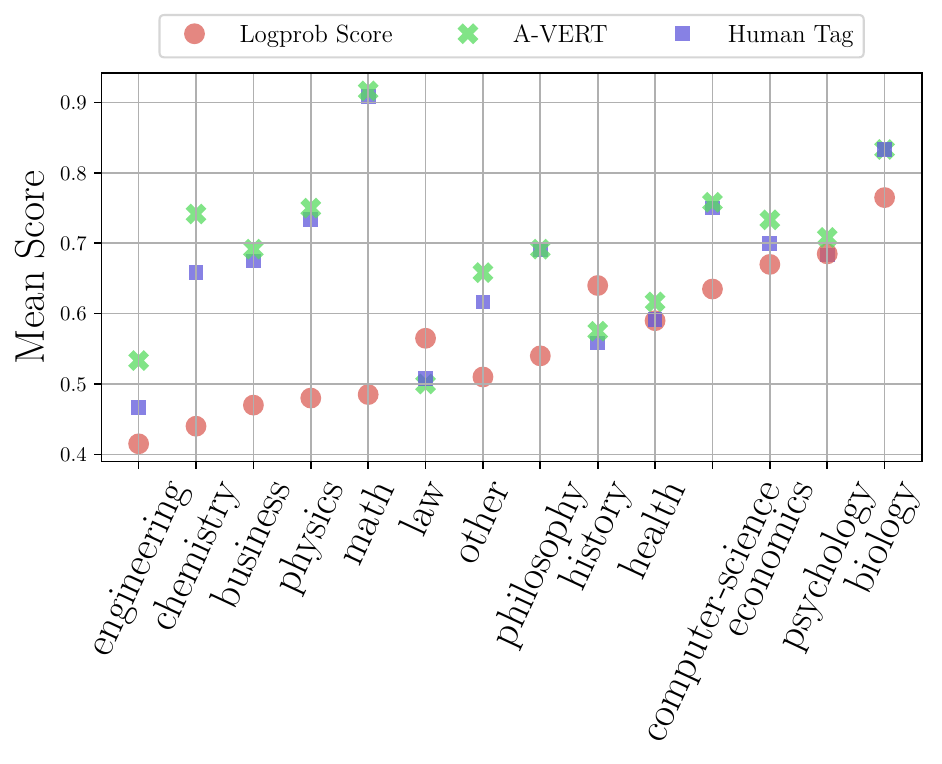}
       \subcaption{}
        \label{fig:mmlu-pro_comparison_task}
    \end{subfigure}
\caption{Comparison of the mean scores in the MMLU-Pro task for each of the analyzed methods: exact match, \emph{A-VERT} and the human tag. \protect\subref*{fig:mmlu-pro_comparison}) MMLU-Pro benchmark mean score using three different scoring techniques. In blue the \emph{{GPT-OSS 20B}} scores, in orange the \emph{Meta Llama 3.3 70B Instruct} scores and in green the \emph{Qwen3 30B A3B} scores. \protect\subref*{fig:mmlu-pro_comparison_task}) Per task scores for the exact match (red circle), \emph{A-VERT} (green cross) and human tag (blue square). The \emph{{GPT-OSS 20B}} results are excluded from this graph to better asses the agreement of the methods.}
\label{fig:mmlu-pro_comparison_full}
\end{figure}

The bAbI results are shown in Figure~\ref{fig:babi_comparison_full}, where each LM average score for the exact match and the \emph{A-VERT} method are compared to the human tag. The same is displayed in Figures~\ref{fig:mmlu_comparison_full} and~\ref{fig:mmlu-pro_comparison_full} for the MMLU and MMLU-Pro benchmarks, respectively.

\begin{figure}[htb!]
\centering
    \begin{subfigure}[b]{0.6\textwidth}
        \centering
        \includegraphics[width=\linewidth]{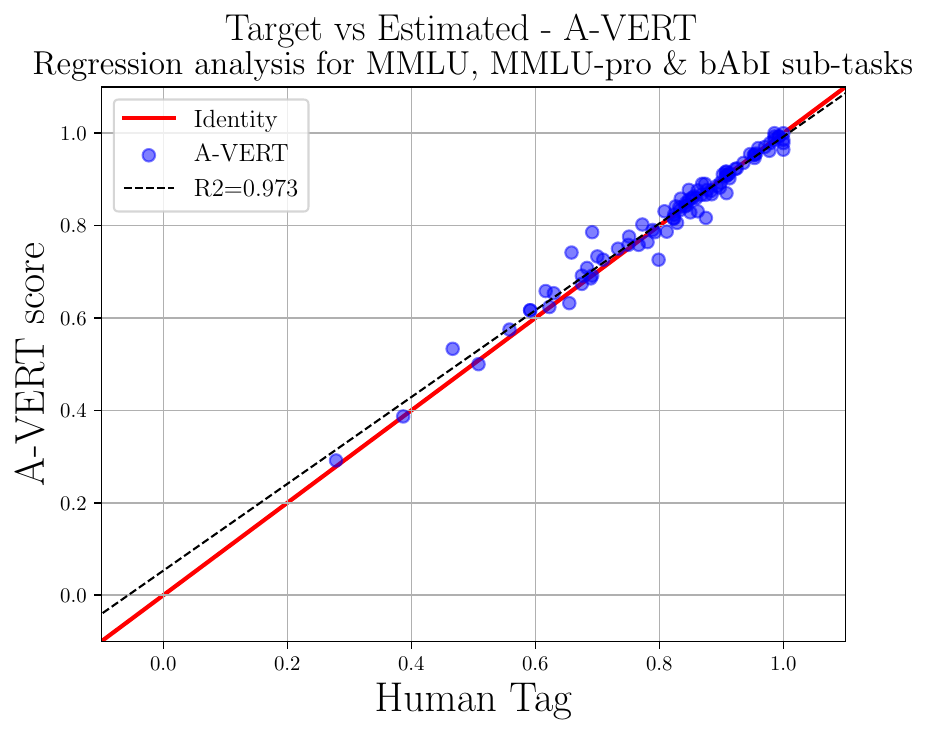}
       \subcaption{}
        \label{fig:overall_regression_avert}
    \end{subfigure}
\caption{\protect\subref*{fig:overall_regression_avert}) Regression of all sub-task scores form bAbI, MMLU and MMLU-Pro using the \emph{A-VERT} against the human annotator scores.}
\label{fig:regression}
\end{figure}

\section{Discussion}\label{sec:discussion}

\subsection{Ranking Models and Ablation Tests}
It can be seen in Figure~\ref{fig:all_embeddings} that the number of the ranking model parameters is low when compared to common \emph{LLM-as-a-Judge} approaches, that tend to use models with hundreds of billion of parameters. In the case of \emph{A-VERT} a very high agreement with human annotators is obtained with models of less than $10 B$ parameters. The ranking models seem to benefit from specialization, this is noted on how the reranker models outperform the embedding models and also on how the instruction enabled models are better when the query is prompted using a problem-specific instruction. In this sense, results of the \emph{A-VERT} methodology are consistent with the limits presented by the embedding models~\cite{wellerTheoreticalLimitationsEmbeddingBased2025}. 
It is also seen how the proposed target enhancements aid the \emph{A-VERT} method, resulting in higher scores every time they are implemented. Finally Table~\ref{table:all_embeddings} highlights that the proposed method does not have a bias towards the \emph{correct} or \emph{wrong} groups, this is observed in the high scores over all presented metrics (for top ranking models).

\subsection{Target Groups Analysis}

The separation observed between the scores for the different quadrants of the confusion matrix (Figure~\ref{fig:all_dist_targets}) show that the \emph{A-VERT} method is capable of providing feedback on the confidence of the ranking function, in other words, the low difference between the \emph{correct} and \emph{wrong} target group scores when the model is making a mistake (\emph{False-Positive/Negative}) and the larger rank difference when the model is correctly selecting the target response (\emph{True-Positive/Negative}) is a very desirable feature that can help discard results based on this metric if higher accuracy is necessary.

It is interesting to note that while a good separation is achieved for the groups both in open-book and multiple-choice questions, the raking scores in the Figure~\ref{fig:babi_dist_targets} are much more separated than in Figure~\ref{fig:mmlu_dist_targets}. This could be the effect of the choices in the MMLU and MMLU-Pro being much more semantically similar between the \emph{correct} and \emph{wrong} groups, a characteristic of multiple-choice exams that seek to confuse the subject under test. Although the method seems to successfully overcome this difficulty, more development should be done to optimize the ranking function in these cases.

\subsection{Benchmarks Test}

When compared to other standard methods such as \emph{exact-match} or \emph{logprobs-score}, the proposed method has a higher correlation with the human annotators, this is summarized in Figure~\ref{fig:regression}, where each sub-task score from Figures~\ref{fig:babi_comparison_full}, \ref{fig:mmlu_comparison_full} and~\ref{fig:mmlu-pro_comparison_full} are plotted against the human tags score. The R2 value of the regression line is $0.967$.
It is important to note that the \emph{A-VERT} method matched the human annotator's response for all models origins, while the other methods failed to obtain results for some LMs. Although the different benchmark configurations used different prompt (\emph{A-VERT} used no instructions nor any examples before the questions, all other methods included at least 3 examples before measuring) we are not analyzing the scores of the LMs, we are interested in provide a consistent evaluation method that can be used in real-world test conditions, in this regard, the proposed method shows great potential. 
Notably, in the task 16 from bAbI, the \emph{exact-match} method signals a higher performance when compared to the \emph{A-VERT} method. This can be explained by the fact that the few-shots provided to the LM in the \emph{exact-match} method provide information that the colors are transferred to the individuals before asking the question; however, with zero-shot, the model lacks the expected skill (that the model knows transitivity). This is an example where few-shot do not only provide structure to the LM response, but also relax the task difficulty (in an uncontrolled manner). On the other hand, in task 19 from bAbI, the opposite is observed: by allowing the model to freely create an answer, its pathfinding ability is better than that measured by \emph{exact-match}. 
This latter finding occurs for most tasks in MMLU.
On the contrary, in the sub-task \emph{Math} from MMLU-Pro, it can be observed that the \emph{logprobs} method, due to the bounded nature of the measurement limiting the number of considered answers, the LM is significantly overpenalized with respect to the same LM but using the \emph{A-VERT} method, where the LM is allowed to generate freely its response.

Regarding the limitations of the \emph{A-VERT} method, it is important to note that the method requires at least one example for each group (\emph{correct} and \emph{wrong} for the current implementation). While this is not an issue for most of the current QA style benchmarks, because they usually provide this in the multiple choices, this could not be the case for future benchmarks of the same type. 
Moreover, while the current method isn't directly applicable to benchmarks such as code generation, summarization, or translation, future research could explore adapting specialized topic embeddings like those proposed for coding \cite{guoGraphCodeBERTPretrainingCode2021} to extend \emph{A-VERT}'s applicability.

\section{Conclusion and Future Work}\label{sec:conclusion}

In this paper we have demonstrated that a semantic scoring method, based on embeddings or rerankers models is able to obtain a high agreement ($>96\%$ accuracy) with human annotators in extracting the semantic belonging of LM responses to a group of targets. As opposed to \emph{LLM-as-a-judge}, the method relies on small models and provides a mathematical expression (a distance) to assess the semantically relevant target group.

\emph{A-VERT} relieves the benchmark creators from the task of adapting (and potentially contaminating) their measurement process to allow them to work for particular LMs. The proposed method enables the response extraction from real-world test conditions, using chat templates and enabling the LM to use any sort of compute-time enhancement in a transparent manner, making the benchmark's results more relatable to day-to-day usage of the LMs.

We expect to build upon the basis of \emph{A-VERT} to create more descriptive scores, for example, the agnostic nature of the method can be used to add more response groups, and capture scenarios where the model refuses to answer or express that the question being answer is faulty (some examples were encountered but no consistent extraction was yet achieved). Also the semantic ranking function (Eq.~\ref{equ:ranker_fn}) can be improved to obtain a baseline semantic similarity using context derived candidate texts and improve the groups separation and achieve a normalized ranking score.

\printbibliography

\end{document}